\pdfoutput=1
\documentclass[11pt]{article}
\usepackage[]{ACL2023}

\usepackage{times}
\usepackage{latexsym}
\usepackage{amsmath,amssymb,amsfonts}
\usepackage[linesnumbered,ruled,vlined]{algorithm2e} 
\usepackage{mathrsfs}
\usepackage{indentfirst}
\usepackage{url}
\usepackage{xspace}
\usepackage{graphicx}
\usepackage{textcomp}
\usepackage{xcolor}
\usepackage{xspace}
\usepackage{subfigure}
\usepackage{multirow}
\usepackage{makecell}
\usepackage{booktabs}
\usepackage{threeparttable}
\usepackage{tablefootnote}
\usepackage{url}
\def\BibTeX{{\rm B\kern-.05em{\sc i\kern-.025em b}\kern-.08em
    T\kern-.1667em\lower.7ex\hbox{E}\kern-.125emX}}
    
\newcommand{\method}{\textbf{\texttt{KnowledgeDA}}\xspace}
\usepackage[T1]{fontenc}
\usepackage[utf8]{inputenc}
\usepackage{microtype}
\usepackage{inconsolata}

\title{A Unified Knowledge Graph Augmentation Service for Boosting Domain-specific NLP Tasks}

\author{Ruiqing Ding$^{1,2}$, Xiao Han$^{*3}$, Leye Wang\thanks{*Corresponding authors} $^{1,2}$\\
$^{1}$Key Lab of High Confidence Software Technologies (Peking University), \\ Ministry of Education, China\\
$^{2}$School of Computer Science, Peking University, Beijing, China\\
$^{3}$School of Information Management and Engineering, \\ Shanghai University of Finance and Economics, Shanghai, China\\
\texttt{ruiqingding@stu.pku.edu.cn, xiaohan@mail.shufe.edu.cn, leyewang@pku.edu.cn}
}

\begin{document}
\maketitle

\begin{abstract}
By focusing the pre-training process on domain-specific corpora, some domain-specific pre-trained language models (PLMs) have achieved state-of-the-art results. 
However, it is under-investigated to design a unified paradigm to inject domain knowledge in the PLM fine-tuning stage.
We propose \method, a \textit{unified} domain language model development service to enhance the task-specific training procedure with domain knowledge graphs.
Given domain-specific task texts input, \method can automatically generate a domain-specific language model following three steps: 
(i) localize domain knowledge entities in texts via an embedding-similarity approach; (ii) generate augmented samples by retrieving replaceable domain entity pairs from two views of both knowledge graph and training data; (iii) select high-quality augmented samples for fine-tuning via confidence-based assessment.
We implement a prototype of \method to learn language models for two domains, \textit{healthcare} and \textit{software development}. Experiments on domain-specific text classification and QA tasks verify the effectiveness and generalizability of \method. 
\end{abstract}

\section{Introduction}
Although general NLP models such as GPT-3 \cite{DBLP:journals/corr/abs-2005-14165} have demonstrated great potential, they may not consistently perform well in domain-specific tasks like healthcare \cite{8440842} and programming \cite{8952330}. 
This is because most pre-trained language models are trained on general-domain corpora, e.g., OpenWebText \cite{radford2019language} and C4 \cite{10.5555/3455716.3455856}. 
However, the words or knowledge entities frequently used in a specific domain are typically different from those in a general domain. 
For instance, scientific texts use different words than general texts, with only a 42\% overlap \cite{Beltagy2019SciBERT}. 
Consequently, general PLMs struggle to capture many important domain entities that rarely appear in general corpora. Therefore, it is necessary to develop a suitable training mechanism for domain-specific NLP tasks.

In general, two steps are needed for domain-specific NLP model development: (i) language model pretraining and (ii) task-specific model training \cite{gu2021domain}. 
Most existing studies focus on pretraining. In particular, to learn domain-specific word embeddings, they retrain PLMs with domain-specific corpora, including ClinicalBERT \cite{alsentzer-etal-2019-publicly}, BioBERT \cite{lee2020biobert}, SciBERT \cite{Beltagy2019SciBERT}, etc. 
In contrast, how to improve the second step (i.e., task-specific training) is under-investigated. 
A common practice is directly fine-tuning the task-specific model with annotated data \cite{gu2021domain}. 
However, \textit{it is difficult to obtain abundant annotated data for a domain-specific task, as labeling often requires domain experts' knowledge}~\cite{DBLP:conf/acl/YueGS20}; without sufficient data, direct fine-tuning may not lead to a satisfactory performance due to overfitting~\cite{pmlr-v126-si20a}. 
Some studies propose task-dependent methods to train task-specific models by introducing some types of domain knowledge \cite{zhu-etal-2022-knowledge}, but they are hard to be generalized to other tasks~\cite{DBLP:conf/icse/TushevEM22}. 

Then, a research question appears: \textbf{can we introduce domain knowledge to task-specific model training in a \textit{unified} way?} 
To answer the question, two main issues need to be addressed: (i) \textit{where to find a unified format of domain knowledge?} (ii) \textit{how to improve the task-specific training of various domains' models in a unified way?}

On one hand, the domain knowledge graph (KG) is an effective and standardized knowledge base for a specific domain \cite{abu2021domain}. 
KGs have been constructed for various domains such as cybersecurity~\cite{jia2018practical}, social-impact funding~\cite{li2020domain}, and healthcare~\cite{li2020real,zhang2020hkgb}, which emphasizes the wide availability of domain KGs. 
Hence, \textit{domain KG could be a feasible source for unified domain knowledge}.
On the other hand, data augmentation (DA) is a data-space approach to enrich training data to avoid overfitting regardless of the task-specific model structure. 
They are often \textit{task-agnostic}~\cite{longpre_how_2020}, i.e., not specified to any particular task. 
This property inspires us that \textit{it may be possible to design a unified DA process to introduce domain knowledge to task-specific model training}. 
However, current DA methods in NLP are mostly proposed for general texts \cite{wei_eda_2019}, and the performance on domain-specific tasks is limited \cite{feng_survey_2021}.
In general, domain-specific DA is still an under-researched direction \cite{feng_survey_2021}.

To fill this research gap, by exploiting domain KGs, we propose \textbf{\method}, a novel and unified three-step procedure to perform domain-specific DA: 
(i) \textit{domain knowledge localization} to map phrases in the text to entities in the domain KG; 
(ii) \textit{domain knowledge augmentation} to fully utilize the KG and the training data to achieve domain-specific augmentation; 
and (iii) \textit{augmentation quality assessment} to single out high-quality augmented data for fine-tuning the task-specific model. 
Specifically:
(i) To the best of our knowledge, this is one of the pioneering efforts toward proposing a unified development process for domain-specific NLP models, especially focusing on task-specific model training.
(ii) \method consists of three core steps, \textit{domain knowledge localization}, \textit{domain knowledge augmentation}, and \textit{augmentation quality assessment}. 
We implement a prototype of \method, which can automatically learn domain-specific models given domain-specific texts, especially in \textit{healthcare} domain.
(iii) Experiments are run on text classification and QA tasks (English and Chinese) mainly in healthcare. Results show that \method can obtain $\sim 4\%$ improvement compared to direct fine-tuning, and significantly outperform existing DA methods \cite{wei_eda_2019,yue_phicon_2020}.
The source codes are available\footnote{\scriptsize{https://github.com/RuiqingDing/KnowledgeDA}}.

\section{Related Work}
\noindent \textbf{Domain-specific Knowledge-augmented NLP Methods}.
To improve domain-specific NLP model development, a general strategy is introducing domain knowledge \cite{zhu-etal-2022-knowledge}. 
For zero and few-shot text classification tasks, KPT \cite{hu-etal-2022-knowledgeable} incorporates external knowledge into the projection between a label space and a label word space. 
For text generation, KG-BART \cite{liu2021kg} proposes a novel knowledge graph augmented pre-trained language generation model to promote the ability of commonsense reasoning.
For question answering and dialogue, some work use external knowledge bases to inject commonsense, like KaFSP \cite{li-xiong-2022-kafsp}, KG-FiD \cite{yu-etal-2022-kg}, etc. 
Besides task-dependent methods, there are also some \textit{unified} training strategies to incorporate knowledge (domain-specific corpora) into PLMs, leading to domain-specific PLMs such as BioBERT~\cite{lee2020biobert}, SciBERT \cite{Beltagy2019SciBERT}, ClinicalBERT \cite{alsentzer-etal-2019-publicly}, and UmlsBERT~\cite{michalopoulos-etal-2021-umlsbert}.
Also, there are three primary techniques to integrate knowledge graphs and PLMs: 
(i) pre-training a PLM from scratch by using KG or other structural knowledge/texts \cite{feng2022kalm, huang2022endowing}; 
(ii) adapting a given PLM to incorporate KG information with new network layers in task-specific training/fine-turning \cite{zhang2022greaselm,yasunaga2022deep,kang-etal-2022-kala};
(iii) augmenting training data with KGs during task-specific training/fine-tuning, e.g., PHICON \cite{yue_phicon_2020}.
Our work also attempts to improve the domain-specific NLP model development in a unified manner. Different from PLM, we focus on task-specific NLP model fine-tuning \cite{gu2021domain}.
Hence, our proposed \method can be used with domain-specific PLMs together to construct NLP models.

\noindent \textbf{Text Data Augmentation (DA)}.
DA has received increasing interest, especially low-resource situations\cite{feng_survey_2021}. 
In general, there are three types of text DA methods:
(i) \textit{Rule-based} techniques, e.g., EDA \cite{wei_eda_2019}, adopt token-level random perturbation operations including random insertion, deletion, and swap; 
(ii) \textit{Interpolation-based} techniques, pioneered by MIXUP \cite{DBLP:conf/iclr/ZhangCDL18}, interpolate the inputs and labels of two or more real examples. Follow-ups include SwitchOut \cite{DBLP:conf/emnlp/WangPDN18}, MixText \cite{chen-etal-2020-mixtext}, etc; 
(iii) \textit{Generator-based} techniques, e.g., LAMBADA \cite{DBLP:conf/aaai/Anaby-TavorCGKK20} and GPT3Mix \cite{DBLP:conf/emnlp/YooPKLP21}, learn generators by fine-tuning the large language generation models (e.g., GPT) on the training data to generate new samples. 
Basically, three types of methods can be used together as they augment data from diverse perspectives.
However, regardless of the type, most existing studies do not explicitly introduce domain knowledge. 
PHICON \cite{yue_phicon_2020} attempts to use the domain-entity dictionary for text DA, which replaces an entity mention in a sentence with another same-category entity. 
Compared to PHICON, \method further considers relationships in the domain KG; besides, \method introduces other newly-designed components, e.g., augmentation quality assessment, to ensure high-quality augmentation.

\begin{figure*}
    \centering
    \includegraphics[width=0.9\linewidth, height=0.4\linewidth]{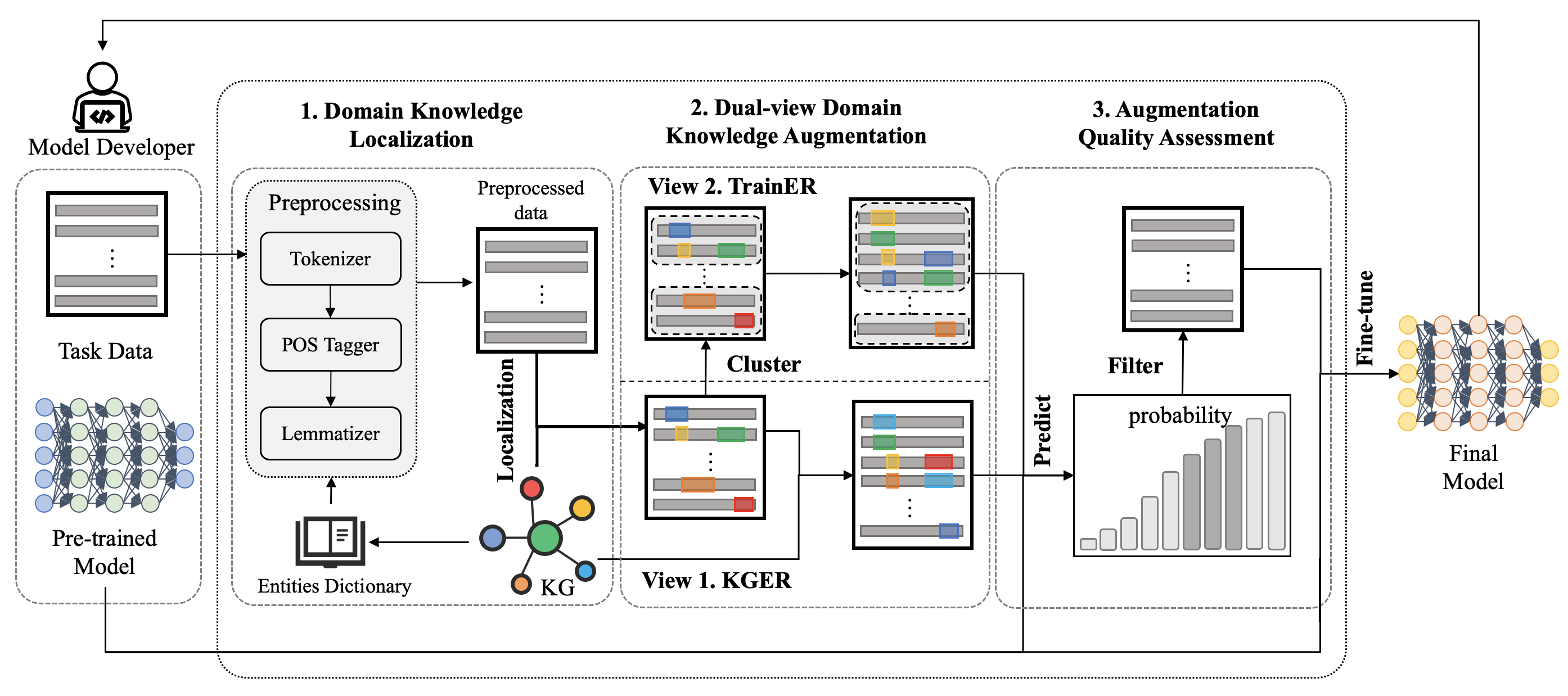}
    \vspace{-1em}
    \caption{\small{Overview of \method: the user only needs to upload the task text data (i.e., training data) and specifies the pre-trained language models (e.g., BERT). All the domain knowledge injection procedures are automatically conducted. Finally, a well-performed domain-specific NLP model will be obtained from \method.}}
    \label{fig: knowledgeDA}
    \vspace{-1em}
\end{figure*}

\section{The \method Framework}

\subsection{Workflow of \method}
To facilitate the development of domain-specific NLP models, we propose a unified domain KG service, \method, 
which can achieve explicit domain knowledge injection by domain-specific DA. 
Challenges to be addressed include:

\textbf{C1. How to discover domain knowledge in texts?}
Detecting entities in a text is the first step to link the text with a knowledge base.
A domain entity may have multiple expressions, e.g., \textit{lungs} and \textit{pulmonary} share a similar meaning in the healthcare domain.
It is important to deal with synonyms.

\textbf{C2. How to ensure that the augmented texts retain the domain information and are semantically correct?}
We aim to achieve interpretable data augmentation through explicit domain knowledge injection. 
The domain information and the semantic correctness of augmented samples are desirable to be kept after data augmentation. 

\textbf{C3. How to ensure the quality of augmented texts?}
As PLMs grow larger, simple DA method becomes less beneficial \cite{feng_survey_2021}. 
It is essential to select beneficial samples from all the augmented samples for efficient fine-tuning.

To address the above challenges, we design corresponding modules in \method (shown in Figure~\ref{fig: knowledgeDA}):
(i) \textbf{\textit{domain knowledge localization}}, which locates the mentions of domain KG entities in texts; 
(ii) \textbf{\textit{domain knowledge augmentation}}, which incorporates a dual-view DA strategy by considering both domain KG and training data; 
(iii) \textbf{\textit{augmentation quality assessment}}, which retains beneficial augmented samples for fine-tuning using a confidence-based strategy.
When the task data and the PLM (e.g., BERT) are given, \method can automatically conduct data augmentation based on built-in domain KGs and output the final domain task-specific model.

\subsection{Module 1: Domain Knowledge Localization}
\label{subsection:module1}

\begin{figure}
    \centering 
    \includegraphics[width=0.85\linewidth]{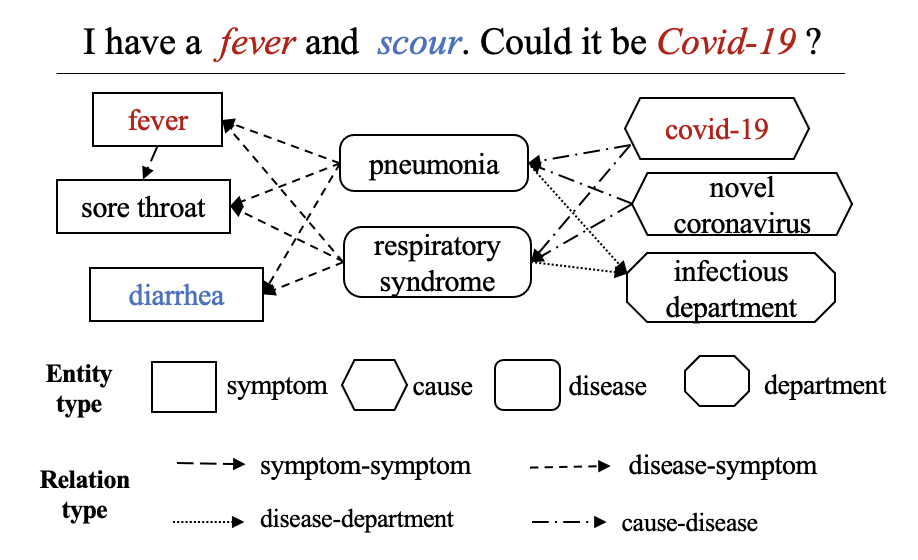}
    \vspace{-1em}
    \caption{\small{An example of domain knowledge localization in the healthcare domain. The words in {\color{red}red} indicate that the entities and the mentions are \textit{exactly} same, and the words in {\color{blue}blue} indicate that the mention (scour) and the entity (diarrhea) share \textit{similar word embeddings}.}}
    \label{fig:localization}
    \vspace{-1em}
\end{figure}

\begin{figure*}
    \centering 
    \includegraphics[width=0.95\textwidth, height=0.4\linewidth]{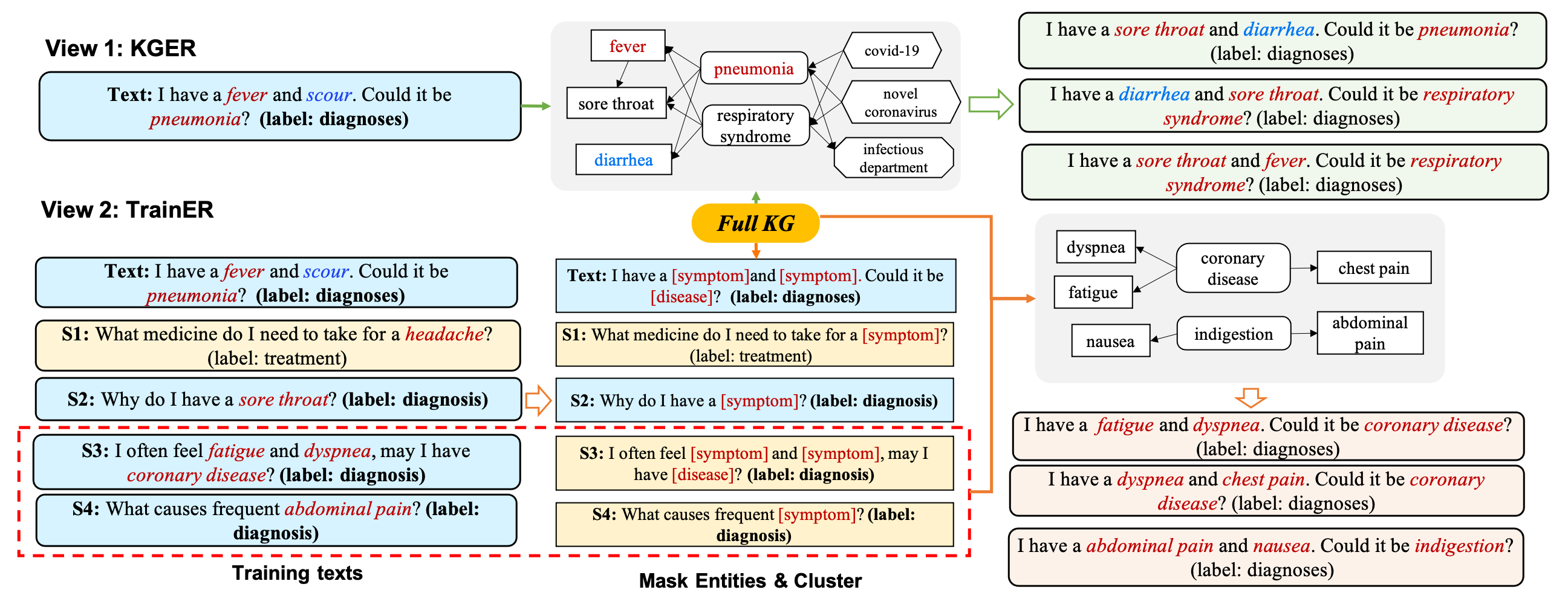}
    \vspace{-1em}
    \caption{\small{Dual-view knowledge augmentation with a healthcare text as an example. \textbf{KGER}: 1. retrieve 2-hop relevant entities in the full KG; 2. replace entities following two principles, `entity relevance' and `relation consistency'. \textbf{TrainER}: 1. mask entity with the category to represent the expressions pattern; 2. text clustering to select samples with the same label but different clusters (i.e., S3 \& S4); 
    3. collect the candidate triples from selected samples; 4. augment data by replacing relevant entities.}}
    \label{fig:augmentation}
    \vspace{-1em}
\end{figure*}

Detecting entities in texts can identify domain-specific objects and the relations between them.
Considering that an entity may correspond to multiple mentions \cite{DBLP:conf/naacl/FlorianHIJKLNR04}, \textit{exact} string matching will lead to a low matching rate.
Although there are some open entity detection tools, like TAGME \cite{DBLP:conf/cikm/FerraginaS10} and BLINK \cite{DBLP:conf/emnlp/WuPJRZ20}, and some studies achieve supervised non-exact matching of entities and mentions  \cite{DBLP:journals/symmetry/HuTZGX19}, the performance on domain-specific entities can not be guaranteed.
Then, we use an annotation-free string-similarity-based strategy \cite{DBLP:conf/eacl/BunescuP06,karadeniz2019linking} to discover \textit{non-exact but correct} mappings between mentions in the text and entities in the KG.
Specifically, we calculate the inner product of word embeddings as string similarity \cite{DBLP:conf/emnlp/WuPJRZ20}.

As seen in Figure~\ref{fig: knowledgeDA}, we follow the NLP preprocessing pipeline and match the processed text with KG.
During preprocessing, we add the entity strings in KG to the dictionary of \textit{tokenizer} to avoid word segmentation errors, e.g., `cerebral embolis' should be treated as a medical term rather than being splitted into two words.
Also, we use \textit{POS Tagger} and \textit{Lemmatizer} to convert each token to the canonical form (lemma), e.g., the lemma for `coughed' is `cough'. 
While knowledge localization, we extract the entities' embeddings and the mentions' embeddings from the PLM, and then calculate the similarity between them.
We consider the pair of a mention and the most similar KG entity as a match if the similarity score is larger than a threshold $\lambda$ (0.9 in our implementation). 

An example in healthcare is illustrated in Figure~\ref{fig:localization}.
Without similarity match, we will ignore that \textit{scour} and \textit{diarrhea} are analogous. 
Through localization, some relations between entities can be also constructed, e.g., fever and scour are symptoms of pneumonia and respiratory syndrome.
These will be used in the next module for data augmentation.

\subsection{Module 2: Dual-view Domain Knowledge Augmentation}
\label{subsection:module2}

After locating the domain knowledge, i.e., entity mentions in the text, the next step is to replace these mentions with other \textit{relevant entity words} for domain-specific data augmentation. 
Here, we propose a dual-view strategy to conduct the \textit{relevant entity retrieval} by considering both KG and the training text data.

\vspace{+.5em}
\noindent \textbf{View 1: KG-based Entity Retrieval (KGER)}

A direct strategy for domain KG-based DA is to replace the entity with another same-category entity, e.g., replacing `William' with `Mike' as both are person names \cite{yue_phicon_2020}.
However, it may suffer from two pitfalls:
(i) Although the original and replaced entities are in the same category, they can be totally different, such as \textit{pneumonia} and \textit{fracture} (both are diseases), which may negatively impact the downstream tasks, e.g., classifying a medical transcription to the relevant department\footnote{\scriptsize{https://www.kaggle.com/datasets/tboyle10/medicaltranscriptions}};
(ii) When two or more entities appear in a text, they may have certain valuable relationships (e.g., disease and symptom), but replacing these entities separately would ignore this information.

To address the above issues, we propose two principles for \textbf{KGER}: 
(i) \textit{entity relevance}, refers to ensuring that the retrieved entity is similar to the original entity, not just with the same category; 
(ii) \textit{relation consistency}, means keeping the relationships unchanged between multiple replaced entities in one text.
We formulate a domain KG as $\mathcal{G} = \{E, R, T, C\}$, where $E$, $R$, $T$, and $C$ are the sets of entities, relations, triples, and entities' categories, respectively.
Specifically, $T = T^R \bigcup T^C$, where $T^R = \{(h, r, t)| h, t \in E, r \in R \}$ and $T^C = \{(e, BelongTo, c) | e \in E, c \in C\}$.

Given an entity $e$, we can get its category $c$, involved triples $T_e = \{(e, r, t) \in T^R\} \bigcup \{(h, r, e) \in T^R\}$,
and the adjacent entities $E_e = \{e' | (e, r, e') \in T_e, (e, r, e') \in T_e\}$.
To obtain more same-category entities, we further retrieve the involved triples of $E_e$, named $T_{e2}$, and put $T_e$ and $T_{e2}$ together as the candidate triples (i.e., 2-hop triples around $e$).

That is: (1) If only one entity exists, or multiple entities exist but do not have direct KG relations in the text, we randomly select a same-category entity $e'$ from the candidate triples to replace each original entity $e$. Note that $e'$ must be within 2-hop around $e$, ensuring the \textit{entity relevance}. 
(2) If there exist certain pairs of entities with KG relations, we would seek the same relation-type triple from the candidate triples for replacing the pair of entities together, following the \textit{relation consistency}.

For instance, `\textit{I have a \textbf{fever} and \textbf{scour}. Could it be \textbf{pneumonia}?}' (shown in Figure ~\ref{fig:augmentation}), \textit{fever} and \textit{scour} are the symptoms of \textit{pneumonia}. 
So we need to search for suitable triples to satisfy relation consistency. For instance, \textit{diarrhea} and \textit{sore throat} are the symptoms of \textit{respiratory syndrome}, so the augmented text could be `\textit{I have a \textbf{diarrhea} and \textbf{sore throat}. Could it be \textbf{respiratory syndrome}?}'.

\vspace{+.5em}
\noindent \textbf{View 2: Training Data-based Entity Retrieval (TrainER)}

In View 1, we mainly retrieve relevant entities that are close in the KG. 
However, entity pairs far away in the KG may be helpful for the specific task if being replaced with each other. 
For example, for the task to detect the medical query intent, `\textit{blood routine examination}' and `\textit{CT}' is the entity pair that could be replaced with each other for augmentation because they most probably appear in the queries about \textit{diagnosis} and \textit{cause analysis}, but they are distant from each other in the medical KG, like CMedicalKG \footnote{\scriptsize{https://github.com/liuhuanyong/QASystemOnMedicalKG}}.

To find such task-specific valuable replacement entity pairs which may not be near in the KG, we design a new View 2, \textit{Training Data-based Entity Retrieval (TrainER)}, to retrieve task-specific entity pairs from training data. 
REINA \cite{DBLP:conf/acl/WangXFLSX0022} has verified that retrieving from training data to enrich model inputs (concatenating the original input and retrieved training data) may generate significant gains.
Inspired by this idea, TrainER aims to extract gainful entity pairs from the training data for augmentation.

In general, a good entity pair for replacement may satisfy at least two properties: (i) \textit{label consistency}, indicates that the two entities in the pair should be contained in two training texts with the same task label; (ii) \textit{expression diversity}, means that the two texts containing the two entities should have different expression patterns, so as to enrich the training data diversity. 
Specifically, to reach \textit{label consistency}, for an entity $e$ in a text $t$, we would retrieve a same-category entity $e'$ from another text $t'$ if $t$ and $t'$ have the same label. 
To achieve \textit{expression diversity}, we first cluster all the training texts into different clusters with diverse expression patterns. 
Then, for an entity $e$ in a text $t$,  the replaced entity $e'$ will be retrieved from $t'$ only if $t$ and $t'$ are not in the same cluster.  
Figure~\ref{fig:augmentation} elaborates on the process of TrainER. 

To conduct training data clustering to differentiate expression patterns,  
we first mask entities with their categories to extract the expression templates for each training text.
For instance, `\textit{what is \textbf{pneumonia}?}' and `\textit{what is \textbf{fracture}}' share the same expression template `\textit{what is \textbf{[disease]}}?', as both sentences have the same pattern regardless of the specific entity (i.e., disease).
Then, we run a clustering algorithm on the masked texts, i.e., expression templates, to identify diverse expression patterns. 
The K-means clustering \cite{DBLP:conf/soda/ArthurV07} is applied due to its high efficiency and effectiveness in empirical experiments; the feature of a masked text is represented by TF-IDF vectorization \cite{DBLP:journals/jd/Jones04}.
Same as the \textit{relation consistency} principle in KGER, if there are certain entity pair with KG relations in the original text, we will retrieve the entity pair with the same relation from other training texts.

\subsection{Module 3: Augmentation Quality Assessment}
\label{section: module3}
After Module 1 \& 2, we obtain a set of augmented texts. A straightforward way is to fine-tune task-specific models with these texts like most prior studies~\cite{zhang_character-level_2015,wei_eda_2019}. 
Recent work \cite{DBLP:conf/acl/ZhouZTJY22} has found that not all the augmented texts are equivalently effective; thus, selecting high-quality ones may further improve the model performance. 

Inspired by this finding, \method includes a quality assessment module to justify the quality of each augmented text. 
Prior work \cite{DBLP:conf/aaai/Anaby-TavorCGKK20,DBLP:conf/acl/ZhouZTJY22} uses the prediction confidence as the quality metric and selects top-$K$ high-confidence augmented samples for fine-tuning, because this ensures the label correctness of augmented texts.
However, we argue that it may not significantly improve the model performance since high confidence means that the pattern inside the augmented sample has already been encoded in the original model (without augmentation). 

Hence, we first fine-tune PLM (e.g., BERT) on the task texts; then use this plain fine-tuned model $\mathcal M$ to predict the augmented texts and obtain the confidence scores.
Instead of selecting top-$K$ confident samples, we pick $K$ augmented samples whose confidence is close to a predefined threshold $\delta$. 
Note that $\delta$ should not be a too small number, as we still want to ensure the correctness of the training labels for augmented texts; meanwhile, $\delta$ should not be too large, as a very high-confident sample would contribute little new knowledge to the model. 
Based on this idea, we design a novel confidence-based data filtering strategy to retain gainful augmented samples.

The task data $D=\{(x_i, y_i)\}_{i=1}^{n}$ and the plain fine-tuned model $\mathcal{M}$ (without augmentation) are known, where $x_i$ is a string of text, and the label $y_i \in \{1,2, \cdots, q\}$ is the label of $x_i$ among a set of $q$ labels.
Through \textit{KGER} and \textit{TrainER}, we can generate the augmented samples $D_{i}^{aug} = \{x_{i}^{1}, x_{i}^{2}, \cdots, x_{i}^{m}\}$ for the i-th sample, $x_i$. 
The prediction confidence (probability) of $D_{i}^{aug}$ can be calculated as $P_{i}^{aug} = \{p_{i}^{j}\}_{j=1}^{m}$, where $p_{i}^{j} = prob(\mathcal{M}(x_{i}^{j}) = y_i)$.

We propose a confidence threshold $\delta$ to adjust sample selection criteria. 
Given $\delta$, the sampling weights of $D_{i}^{aug}$  can be calculated by
\begin{small}
\begin{equation}
    w_{i}^{1}, w_{i}^{2}, \cdots, w_{i}^{m} = softmax(\xi_{i}^{1}, \xi_{i}^{2}, \cdots, \xi_{i}^{m})
    \label{eq:sampling}
\end{equation}
\end{small}
where $\xi_{i}^{j} = 1-|\delta-p_{i}^{j}|$.
If $p_{i}^{j}$ is closer to $\delta$ (0.75 in our implementation), we have a higher probability to select this sample.
With this confidence-based sampling strategy, we can select augmented samples to further fine-tune the task model $\mathcal M$. In general, the selected samples would be relatively confident but not too highly-confident, thus ensuring both \textit{label correctness} and \textit{new knowledge}.

\renewcommand{\arraystretch}{1.3}
\begin{table}[t]
\footnotesize
    \centering
    \resizebox{\columnwidth}{!}{%
    \begin{tabular}{lcccc}
    \toprule
     \textbf{Dataset} & \textbf{Lang.} & \#\textbf{Labels} & \#\textbf{Samples} & \#\textbf{Mentions}  \\ \hline
     CMID & CHI & 4 & 12254 & 5182 \\
                                 KUAKE-QIC & CHI & 11 & 8886 & 3369 \\
                                 TRANS & ENG & 7 & 1740 & 2298 \\
                                 ABS & ENG & 5 & 14438 & 3808 \\ \hline
    \end{tabular}%
    }
    \vspace{-0.5em}
    \caption{Dataset Statistics}
    \label{tab:data}
    \vspace{-0.5em}
\end{table}

\renewcommand{\arraystretch}{1.3}
\begin{table*}[t]
    \centering
    \scriptsize
    \begin{tabular}{lcccccccccccc}
    \toprule
    \multicolumn{1}{c}{\multirow{2}{*}{\textbf{DA Method}}} & \multicolumn{2}{c}{\textbf{CMID} (Chinese)} & \multicolumn{2}{c}{\textbf{KUAKE-QIC} (Chinese)} & \multicolumn{2}{c}{\textbf{TRANS} (English)} & \multicolumn{2}{c}{\textbf{ABS} (English)} \\ 
    \cmidrule(r){2-3} \cmidrule(r){4-5} \cmidrule(r){6-7} \cmidrule(r){8-9} 
    \multicolumn{1}{c}{} & Acc. & F1 & Acc. & F1 & Acc. & F1 & Acc. & F1  \\ 
    \hline
    \textit{None} & 70.25(0.80) & 68.21(0.90) & 78.82(0.81) & 78.57(0.72) & 73.10(1.79) & 71.50(1.77) & 63.95(0.31) & 62.84(0.40) \\
    \textit{SR} & 71.90(0.76) & 70.97(0.39) & 80.52(0.73) & 80.10(0.80) & 72.38(0.24) & 72.52(0.30) & 64.14(0.25) & 63.13(0.30) \\
    \textit{EDA} & 70.59(0.65) & 70.05(1.25) & 79.45(0.33) & 79.01(0.41) & 73.91(0.26) & 73.71(0.28) & 63.23(0.65) & 62.09(0.72) \\
    \textit{PHICON} & 71.95(0.35) & 71.14(0.53) & 80.52(0.74) & 80.23(0.82) & 74.53(1.19) & 73.10(0.77) & 64.17(0.59) & 63.35(0.60) \\
    \textit{\method} & \textbf{72.38(0.46)*} & \textbf{71.94(0.38)} &
                           \textbf{81.67(0.41)***} & \textbf{81.31(0.44)*} & 
                           \textbf{75.66(0.58)**} & \textbf{75.37(0.72)} & 
                           \textbf{64.97(0.29)} & \textbf{64.18(0.28)*} \\ 
    \bottomrule
    \end{tabular}
    \vspace{-0.5em}
    \caption{\small{Performance of baselines and \method with BERT in healthcare: 1. Values in `( )' denote the standard deviation of five repeated experiments' results; 2. \textbf{Bold} denotes the best-performed ones of the task; 3. *, **, *** denote that the t-test significance p-value $<$ 0.1, 0.05, 0.01 when comparing the results of \method and the best baseline.}}
    \label{tab:MainRes_Healthcare}
    \vspace{-1em}
\end{table*}

\section{Empirical Evaluation}

\subsection{Text Classification}
\subsubsection{Setup}

\noindent \textbf{Datasets.} 
We conduct experiments on four datasets in healthcare: CMID\footnote{\scriptsize{https://github.com/ishine/CMID}} and KUAKE-QIC \cite{DBLP:conf/acl/ZhangCBLLSYTXHS22}are in Chinese; 
TRANS\footnote{\scriptsize{https://www.kaggle.com/datasets/tboyle10/medicaltranscriptions}} and ABS\footnote{\scriptsize{https://github.com/PoojaR24/Medical-Text-Classification}} are in English.
The basic information is enumerated in Table \ref{tab:data}.
For Chinese, we use an open-source medical KG, CMedicalKG\footnote{\scriptsize{https://github.com/liuhuanyong/QASystemOnMedicalKG}}; 
for English, we adopt the Unified Medical Language System (UMLS) \cite{DBLP:journals/nar/Bodenreider04} as KG.
The preprocessing of KGs can be found in Appendix \ref{sec:kg_preprocess}.

\vspace{+.5em}
\noindent \textbf{Baselines.} As \method focuses on explicit knowledge injection during DA by domain entity replacement, we mainly compare with state-of-the-art rule-based DA methods: \textbf{SR} \cite{vijayaraghavan-etal-2016-deepstance} uses token-level replacement with synonyms; \textbf{EDA} \cite{wei_eda_2019} uses token-level random perturbation operations including random insertion,  deletion, and swap; \textbf{PHICON} \cite{yue_phicon_2020} uses entity-level replacement with other entities belonging to the same category. 
For each DA method, we scale up the training data to 5 times the original size and select the best model on the validation set for evaluation.
All the methods are based on the same text classifier with the same hyper-parameters (in Appendix \ref{sec:settings}).
In the main experiments, the base classifier is BERT-base (hereinafter referred to as BERT). 
We also experiment with domain-specific PLMs as stronger classifiers, discussed in the later part. 
And the experiment in the software development domain is shown in Appendix \ref{sec:result_software}.

\subsubsection{Results in Healthcare}

\renewcommand{\arraystretch}{1.3}
\begin{table}[t]
\centering
\scriptsize
\begin{tabular}{lcccc}
\toprule
 \textbf{Method} & \multicolumn{1}{l}{\textbf{CMID}} & \multicolumn{1}{l}{\textbf{KUAKE-QIC}} & \multicolumn{1}{l}{\textbf{TRANS}} & \multicolumn{1}{l}{\textbf{ABS}} 
 \\ \hline
\textit{SR} & 14.80\% & 17.89\% & 33.56\% & 24.33\%   \\
\textit{EDA} & 11.18\% & 15.85\% & 26.66\% & 25.07\%  \\
\textit{PHICON} & 35.84\% & 29.11\% & 76.45\% & 74.07\% \\
\textit{\method} & \textbf{40.67\%} & \textbf{35.36\%} & \textbf{79.33\%} & \textbf{78.37\%}
\\ \bottomrule
\end{tabular}
\vspace{-0.5em}
\caption{\small{Novel entity coverage for healthcare datasets}}
\label{tab:diversity}
\vspace{-0.5em}
\end{table}

\begin{table}[t]
    \centering
    \resizebox{\columnwidth}{!}{%
    \begin{tabular}{lcccc}
    \toprule
    \multicolumn{1}{c}{\multirow{2}{*}{\textbf{Method}}} & \multicolumn{2}{c}{\textbf{CMID}} & \multicolumn{2}{c}{\textbf{TRANS}} \\ 
    \cmidrule(r){2-3} \cmidrule(r){4-5}
    \multicolumn{1}{c}{} & Acc. & F1 & Acc. & F1 \\ \hline
    \textit{None} & 73.10(0.32) & 71.71(0.37) & 75.88(0.41) & 75.22(0.57) \\
    \textit{SR} & 73.49(0.19) & 72.08(0.22) & 75.38(1.11) & 75.01(1.43) \\
    \textit{EDA} & 72.93(0.46) & 71.89(0.89) & 75.40(0.79) & 74.91(0.80) \\
    \textit{PHICON} & 73.51(0.68) & 72.49(0.63) & 75.37(0.59) & 75.18(0.72) \\
    \textit{\method} & \textbf{73.60(0.33)} & \textbf{72.61(0.31)} & \textbf{76.52(0.59)**} & \textbf{76.54(0.91)*} \\ \bottomrule
    \end{tabular}%
    }
    \vspace{-0.5em}
    \caption{\small{Performance with domain-specific PLMs, where CMID/TRANS use eHealth/ClinicalBERT as the domain-specific PLM.}}
    \label{tab:domain_plm}
    \vspace{-1em}
\end{table}

\noindent \textbf{Main Results.}
Table \ref{tab:MainRes_Healthcare} shows the results of different DA methods. 
We can observe that \method achieves the best performance among all the methods in both accuracy and F1 score on four datasets.
At the same time, \textbf{PHICON} also outperforms \textbf{SR} and \textbf{EDA} in most cases, verifying the effectiveness of domain-specific knowledge.
Specifically, on two Chinese datasets, CMID and KUAKE-QIC, \method improves the accuracy by 3.03\% and 3.62\%, respectively, over the fine-tuned model without augmentation. Moreover, compared to the best baseline, \textbf{PHICON}, \method's improvements on accuracy are still statistically significant. Similar results are also observed on two English datasets.
In a nutshell, the results suggest that domain-specific entity replacement can facilitate text classification in healthcare. 
Compared to \textbf{PHICON} which only considers entity categories, \method selects entities from dual views and accounts for the KG relations between them, which further improves the quality of the augmented text and thus achieves a better performance. 
To further quantitatively verify that \method can introduce more domain knowledge,
following \citeauthor{DBLP:conf/acl/0003XSHTGJ22} (\citeyear{DBLP:conf/acl/0003XSHTGJ22}), we calculate \textit{Novel Entity Coverage}, the percentage of the \textit{novel} entities in the test data covered by augmented texts (\textit{novel} means not appearing in the training data).
As illustrated in Table~\ref{tab:diversity}, \method has the highest coverage, which also explains the effectiveness.

\vspace{+.5em}
\noindent\textbf{Domain-specific PLMs as the Base Classifiers.}
Domain-specific PLMs contain domain knowledge by pre-training with domain corpus. 
To confirm that \method is still beneficial with a domain-specific PLM, we use eHealth \cite{DBLP:journals/corr/abs-2110-07244} and ClinicalBERT \cite{alsentzer-etal-2019-publicly} as the PLMs for Chinese and English datasets, respectively.
According to Table \ref{tab:domain_plm}, the improvement brought by the domain-specific PLMs is evident (comparing with the results of BERT in Table \ref{tab:MainRes_Healthcare}).
Consistent with the survey \cite{feng_survey_2021}, we discover that when using the domain-specific PLMs, baseline DA methods may not generate an obvious performance improvement and even have a negative effect compared to no-augmentation.
For instance, on TRANS, \textbf{EDA} improves the performance over BERT (increasing F1 score from $71.50$ to $73.71$); while \textbf{EDA} worsens the performance when a domain-specific PLM is used (reducing F1 score from $75.22$ to $74.91$). 
However, even with domain-specific PLMs, \method can still improve the domain NLP task performance consistently. Note that for TRANS, \method is the only DA method with positive improvement (and this improvement is also statistically significant).

\begin{figure}[t]
    \centering
    \includegraphics[width=0.9\linewidth]{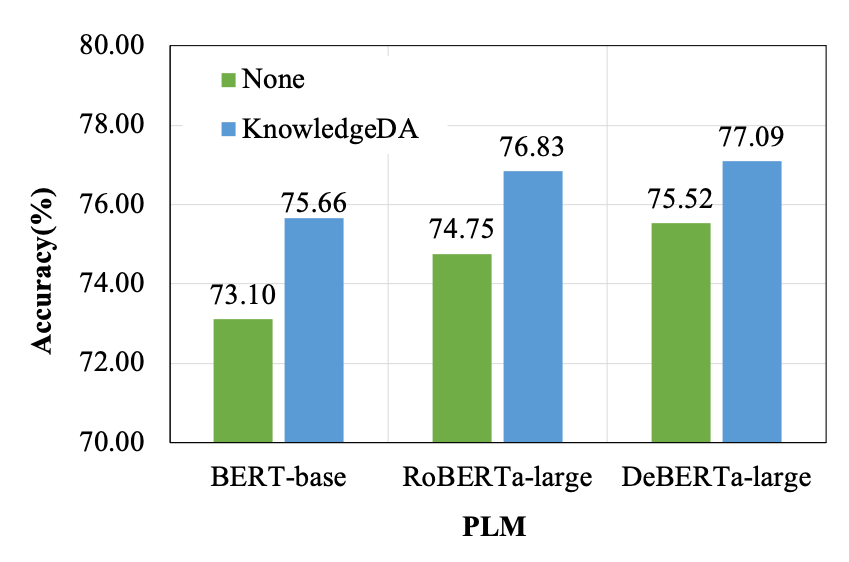}
    \vspace{-1em}
    \caption{Performances on TRANS with larger PLMs.}
    \label{fig:large_plm}
    \vspace{-0.5em}
\end{figure}

\vspace{+.5em}
\noindent\textbf{Larger PLMs as the Base Classifiers.}
In addition to using domain-specific PLMs with the same parameter size as BERT-base (110 
 million parameters), we further take RoBERTa-large \cite{DBLP:journals/corr/abs-1907-11692} and DeBERTa-large \cite{he2021deberta} with more than 350 million parameters as base classifiers on the TRANS dataset.
As shown in Figure \ref{fig:large_plm}, with increasing parameters, RoBERTa-large and DeBERTa-large achieve better accuracy than BERT without DA.
However, there are still notable improvements of 2.78\% and 2.08\% in accuracy with \method on RoBERTa-large and DeBERTa-large, demonstrating the generalizability of \method.

\vspace{+.5em}
\noindent \textbf{Ablation Study.}
To validate the effectiveness of each module of \method, we design corresponding ablation experiments: \textit{KnowledgeDA w.o. SimMatch} removes the similarity-based non-exact matching and only uses exact string matching in Module 1 (Sec.~\ref{subsection:module1}); \textit{KnowledgeDA w.o. KGER} removes the KGER (view 1) in Module 2 (Sec.~\ref{subsection:module2}); \textit{KnowledgeDA w.o. TrainER} removes the TrainER (view 2) in Module 2 (Sec.~\ref{subsection:module2}); \textit{KnowledgeDA w.o. Assess} removes the quality assessment module, i.e., Module 3 (Sec.~\ref{section: module3}).
Table \ref{tab:Res_ablation} shows the results.
\method outperforms all the other methods that remove certain components. 
This verifies the validity of each module of \method.

\begin{table}[t]
    \centering
    \resizebox{\columnwidth}{!}{%
    \begin{tabular}{lcccc}
    \toprule
    \multirow{2}{*}{\textbf{Method}} & \multicolumn{2}{c}{\textbf{CMID}} & \multicolumn{2}{c}{\textbf{TRANS}} \\
    \cmidrule(r){2-3} \cmidrule(r){4-5}
     & Acc. & F1 & Acc. & F1 \\ \hline
    \textit{\method} & \textbf{72.38(0.46)} & \textbf{71.94(0.38)} & \textbf{75.66(0.58)} & \textbf{75.37(0.72)} \\
    \textit{ w.o. SimMatch} & 71.75(0.38) & 71.26(0.41) & 74.05(1.11) & 74.23(0.78) \\
    \textit{ w.o. KGER} & 71.82(0.49) & 71.59(0.35) & 73.92(0.66) & 74.67(0.85) \\
    \textit{ w.o. TrainER} & 71.88(0.57) & 71.48(0.34) & 74.36(0.63) & 74.74(0.65) \\
    \textit{ w.o. Assess} & 72.00(0.61) & 70.84(0.56) & 74.78(0.60) & 74.65(0.73) \\ \bottomrule
    \end{tabular}%
    }
    \vspace{-0.5em}
    \caption{\small{Effectiveness of each module in \method}}
    \label{tab:Res_ablation}
    \vspace{-0.5em}
\end{table}

\begin{table}[t]
\centering
\resizebox{\columnwidth}{!}{%
\begin{tabular}{cccccc}
\toprule
Time(min) & None & SR & EDA & PHICON  & KnowledgeDA\\
\hline
\textit{CMID} & 5.05 & 22.23 & 20.63 & 30.77 & 36.33 \\
\textit{TRANS} & 5.43 & 9.54 & 8.22 & 18.46 & 25.82\\
\bottomrule
\end{tabular}%
}
\vspace{-0.5em}
\caption{\small{Time Consumption of DA \& Model Fine-tuning.}}
\label{tab:time}
\vspace{-0.5em}
\end{table}

\vspace{+.5em}
\noindent \textbf{Time Consumption.}
Table~\ref{tab:time} reports the time consumption of all DA methods on CMID and TRANS. The time required for fine-tuning without augmentation is short ($\sim$ 5 minutes). 
As \textbf{PHICON} and \method need to retrieve the entity mentions and then replace them, time consumption is increased.
In particular, \method takes more time because it considers the relations between entities in the KG. 
In general, the learning process can be completed in about half an hour for \method (also not much longer than competitive baselines like \textbf{PHICON}).

\begin{figure}[t]
\centering
    \begin{minipage}[t]{0.46\linewidth}
        \centering
        \includegraphics[width=1\linewidth]{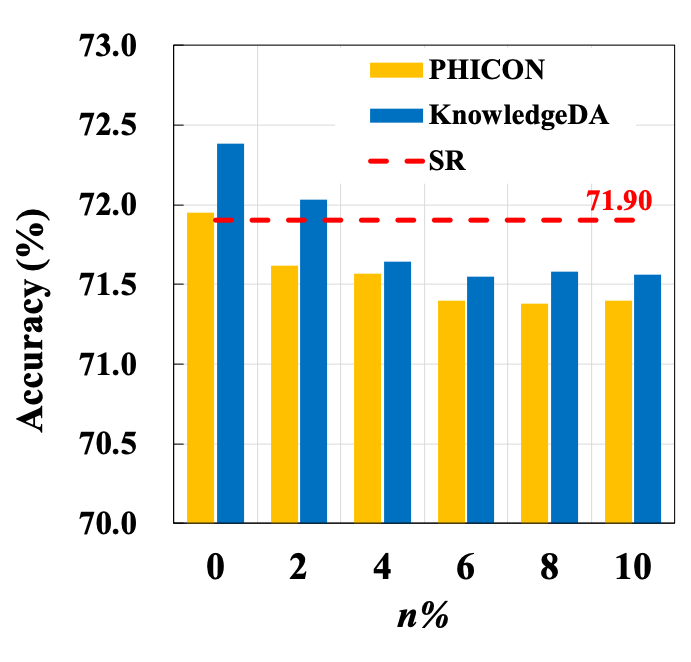}
        \caption{\small{Performance on CMID under n\% CMedicalKG disturbance.}}
        \label{fig:kg_change}
    \end{minipage}
    \begin{minipage}[t]{0.46\linewidth}
        \centering
        \includegraphics[width=1\linewidth]{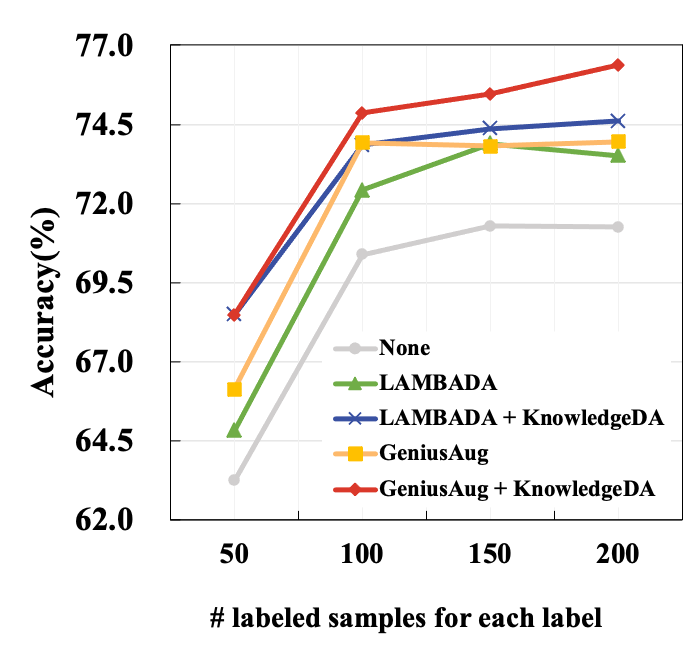}
        \caption{\small{Combining with LAMBADA and GeniusAug (TRANS).}}
        \label{fig:combination}
    \end{minipage}
    \vspace{-1em}
\end{figure}

\vspace{+.5em}
\noindent \textbf{Impact of KG Errors.}
Considering that the KG quality may affect the quality of the augmented texts \cite{kang-etal-2022-kala}, we randomly change the categories of \textit{n}\% entities and the relation types of \textit{n}\% triples in CMedicalKG, and test the performance of \method in CMID dataset.
As shown in Figure~\ref{fig:kg_change}, when we adjust \textit{n} from 0 to 10, the accuracy is between 71.0 and 72.5, with a slight decline. 
When \textit{n} $\geq$ 4, \textbf{SR}, the KG-independent DA method,  performs better. This illustrates the importance to ensure the KG quality for KG-based DA methods, which is consistent with the findings of other KG-based applications \cite{hu-etal-2022-knowledgeable}. In the future, we will explore how to identify potential KG errors so as to improve the robustness of \method.

\vspace{+.5em}
\noindent \textbf{Different Strategies for Augmented Data Quality Assessment and Selection.}
We compare two strategies for augmented data quality assessment and selection: \textit{$\delta$-K} is proposed in Sec.~\ref{section: module3}; \textit{Top-K} \cite{DBLP:conf/aaai/Anaby-TavorCGKK20,DBLP:conf/acl/ZhouZTJY22} selects the top $K$ augmented samples with the highest confidence for each original sample.
Table \ref{tab:Res_assess} shows the results of different strategies, as well as the results of \textit{KnowledgeDA without quality assessment}.
\textit{$\delta$-K} and \textit{Top-K} both outperform \textit{KnowledgeDA without assessment}, verifying the necessity to select high-quality samples for augmentation.
And \textit{$\delta$-K} performs better than \textit{Top-K}.
This empirically validates our intuition that an augmented sample with a not-too-high confidence may bring more new knowledge to the NLP model, as discussed in Sec.~\ref{section: module3}.

\begin{table}[t]
    \centering
    \resizebox{\columnwidth}{!}{%
    \begin{tabular}{lcccc}
    \toprule
    \multirow{2}{*}{\textbf{Method}} & \multicolumn{2}{c}{\textbf{CMID}} & \multicolumn{2}{c}{\textbf{TRANS}} \\
    \cmidrule(r){2-3} \cmidrule(r){4-5}
     & Acc. & F1 & Acc. & F1 \\ \hline
    \textit{w.o. Assess} & 72.00(0.61) & 70.84(0.56) & 74.78(0.60) & 74.65(0.73) \\
    \textit{Top-K} & 71.95(0.52) & 71.53(0.43) & 75.01(0.81) & 74.89(0.69) \\ 
    \textit{$\delta$-K} & \textbf{72.38(0.46)} & \textbf{71.94(0.38)} & \textbf{75.66(0.58)} & \textbf{75.37(0.72)} \\
    \bottomrule
    \end{tabular}%
    }
    \vspace{-0.5em}
    \caption{\small{Different quality assessment strategies in \method}}
    \label{tab:Res_assess}
    \vspace{-1em}
\end{table}

\vspace{+.5em}
\noindent \textbf{Combine with Generator-based Augmentation Techniques.}
\method provides a unified framework for domain-specific knowledge augmentation, which may be combined with other DA techniques. 
Here, we use generator-based augmentation methods as an example. 
Specifically, we generate augmented samples with two methods, LAMBADA \cite{DBLP:conf/aaai/Anaby-TavorCGKK20} and GeniusAug \cite{genius2022guo}; based on these augmented samples, we leverage \method to acquire more augmented samples.
Since generator-based methods are mostly applied to few-shot tasks \cite{DBLP:conf/aaai/Anaby-TavorCGKK20}, we randomly select 50 to 200 samples for each task label in the TRANS dataset. 
LAMBADA and GeniusAug both generate 200 more samples for each label. 
Figure~\ref{fig:combination} shows the results. As expected, the performance of each method goes up as the number of labeled samples increases.
More importantly, combining \method with LAMBADA or GeniusAug both can achieve higher accuracy. 
This demonstrates the general utility of \method to combine with generator-based DA methods to improve the few-shot NLP tasks.

\vspace{+.5em}
\noindent \textbf{Compare with GPT-3.5.}
Recently, ChatGPT has shown powerful text generation capabilities.
To explore the performance of this large language model on domain-specific tasks, we use the OpenAI API \footnote{https://platform.openai.com} to query text-davinci-003 (the most powerful GPT-3.5) by the prompt, \textit{‘decide which label the following text belongs to, \{label names\}: $\backslash$n Text:\{sentence\} $\backslash$n Label: '}. 
It can be seen as a zero-shot manner to response directly.
For TRANS(English), the test accuracy is 66.67\% ($\sim10\%$ lower than \method with BERT). 
It performs even worse on CMID(Chinese) with an accuracy of only 32.32\%, perhaps due to the limited exposure to relevant texts and knowledge.
Therefore, more effective prompt engineering or fine-tuning of GPT (especially for non-English languages) is still necessary for domain-specific tasks, which may be potential future work.

\subsection{QA Tasks}
\noindent \textbf{Setup.}
The \textbf{CMedQA}(Chinese) \cite{zhang2017chinese} and \textbf{PubMedQA}(English) \cite{jin2019pubmedqa} are used for the QA task. 
Both datasets give the label of each question-answer pair (i.e., match or mismatch). 
For CMedQA, we sample 1000 question-answer pairs from the original dataset. 
For PubMedQA, we keep the original data size (429 samples).
In \method, we take the question and answer pair as input and retrieve the entity mentions together.
While fine-tuning, we feed questions and answers, separated by the special [SEP] token to BERT \cite{jin2019pubmedqa}.
The KGs and other settings are the same as classification tasks.

\vspace{+.5em}
\noindent \textbf{Results.}
Table \ref{tab:qa_result} compares the performance of different DA methods based on BERT. 
It is obvious that using any data augmentation strategy can make the performance more stable under different seeds (i.e. smaller standard deviation).
Also, \method outperforms all the baselines.

\begin{table}[t]
    \centering
    \resizebox{\columnwidth}{!}{%
    \begin{tabular}{lcccc}
    \toprule
    \multicolumn{1}{c}{\multirow{2}{*}{\textbf{Method}}} & \multicolumn{2}{c}{\textbf{CMedQA}(Chinese)} & \multicolumn{2}{c}{\textbf{PubMedQA}(English)} \\ 
    \cmidrule(r){2-3} \cmidrule(r){4-5}
    \multicolumn{1}{c}{} & Acc. & F1 & Acc. & F1 \\ \hline
    \textit{None} & 85.00(3.96) & 82.60(7.06) & 66.00(6.87) & 57.65(10.46) \\
    \textit{SR} & 88.46(0.84) & 87.91(0.73) & 72.68(1.97) & 68.99(1.47)\\
    \textit{EDA} & 88.66(1.18) & 88.37(1.00) & 72.72(1.57) & 68.69(1.71)\\
    \textit{PHICON} &  88.56(1.17) & 87.83(1.39) & 73.96(1.88) & 69.67(1.98) \\
    \textit{\method} &  \textbf{89.16(0.58)} & \textbf{88.58(0.56)} & \textbf{74.64(0.83)*} & \textbf{70.98(0.70)}\\ \bottomrule
    \end{tabular}%
    }
    \vspace{-0.5em}
    \caption{\small{Performance of QA Tasks.}}
    \label{tab:qa_result}
    \vspace{-1em}
\end{table}

\section{Conclusions}
\label{sec:future_work}
In this paper, we present \method, a unified knowledge graph service to boost domain-specific NLP tasks.
The intrinsic technical novelty is a three-step framework of task-specific data augmentation process based on domain KGs. 
The experiments on healthcare-related texts both in English and Chinese verify the effectiveness and generality of \method. 
We also confirm that it can be flexible and effective to incorporate other generator-based DA methods on few-shot tasks.
In the future, we can further investigate how to better combine \method and generator-based DA methods and add KG quality inspection methods to avoid the negative impact of errors in KG.

\section*{Limitations}
Domain KGs are the premise of \method, while open and high-quality domain KGs may be rare in some domains.
Therefore, the method will be limited in the domains without suitable KGs.
Besides, we use a similarity-based method to map entity mentions in the text to the corresponding entities in the KG. 
Although this method performs efficiently, it ignores the problem of entity ambiguity \cite{10.1145/3448016.3457328}.
For instance, the abbreviation, CAT, can stand for `\textit{catalase}' or `\textit{COPD Assessment Test}' in healthcare. 
To address this problem, it is necessary to use contextual information to clarify the specific meaning of the mention \cite{DBLP:conf/cikm/PhanSTHL17,DBLP:conf/cidr/OrrLG0ALR21,10.1145/3448016.3457328}.
Last but not least, \method may be not good at tasks of paragraph-level texts and the efficiency will reduce.
Because long texts probably contain more entity mentions and have more complex syntax,  it is more difficult to retrieve the entities and acquire their relations from the KG.

\section*{Ethics Statement}
This paper proposes a unified framework, \method, for text augmentation based on domain KGs for domain-specific NLP tasks.
All the experimental datasets and KGs are publicly available, and the related papers and links have been listed in the paper.
Also, though PLM and KG are publicly available, there may still be several ethical considerations to bear in mind when applying \method in real-world scenarios. For instance, it is crucial to check whether KG contains biases concerning race, gender, and other demographic attributes.

\section*{Acknowledgements}
This research was supported by National Key R\&D Program of China (2020AAA0109401) and NSFC Grants no. 61972008, 72071125, and 72031001.

\bibliographystyle{acl_natbib}
\bibliography{ref}

\clearpage
\appendix

\section{Implementation Details}

\subsection{Experiment Platform \& Settings}
\label{sec:settings}
Our experiment platform is a server with AMD Ryzen 9 3900X 12-Core Processor, 64 GB RAM and GeForce RTX 3090. We use Python 3.6 with pytorch 1.8 on Ubuntu 20.04 for algorithm implementation.

For the text classification task, we feed the [CLS] representation into the output layer when BERT-base as the classifier \cite{devlin_bert_2019}. 
We split the dataset into training set, validation set, and test set as 8:1:1. 
When fine-tuning PLMs, we set batch size to 32, learning rate to 1e-5, and training epoch to 10. 
It will early stop if the loss of the validation set does not decrease in 500 iterations.
\textit{Accuracy} and \textit{micro-F1} are used as the metrics in text classification and QA tasks.
We repeat each experiment 5 times and record the average results.

\subsection{Algorithms}
In this part, we summarize the detailed implementations of \textit{domain knowledge localization} (i.e., Module 1) and \textit{domain knowledge augmentation \& augmentation quality assessment} in Algorithm~\ref{algorithm:localization} and Algorithm~\ref{algorithm:augmentation}, respectively.

\begin{algorithm}[h]
\footnotesize
    \SetAlgoLined 
    \caption{\small{Domain Knowledge Localization}}
    \label{algorithm:localization}
    \KwIn{A text $x$, the entities list $E$, words embeddings dictionary $Embeds$, and similarity threshold $\theta$}
    \KwOut{A matched pair list $Matchs$ of mentions in $x$ and entities in $E$}
    Initialize $Matchs$ as an empty list \;
    Preprocess $x$ with NLP preprocessing pipeline to get word list $words$ \;
    Construct entity embedding matrix $E_{emb}$ and embedding matrix $W_{emb}$ by searching for $E$ and $words$ from $Embeds$ \;
    Compute similarity matrix $Sim = W_{emb} \times E_{emb}.T$ \;
    Query the maximum similarity $sim\_values$ between each word and entity\;
    \If{$sim\_value \geqslant \theta$}{
        Find the index of $sim\_values$ in $Sim$ \;
        Get the pair $(word, entity)$ according to index \;
        Add $(word, entity)$ to $Matchs$ \;
    }
    Return $Matchs$ \;
\end{algorithm}

\begin{algorithm}[h]
    \footnotesize
    \SetAlgoLined 
    \caption{\small{Domain Knowledge Augmentation \& Augmentation Quality Assessment}}
    \label{algorithm:augmentation}
    \KwIn{Train Data $D=\{(x_i, y_i)_{i=1}^{n}\}$; the KG $\mathcal{G} = \{E, R, T, C\}$; the pre-trained language model $PLM$;
    a confidence threshold $\delta$.}
    \KwOut{The selected augmented samples $D^{aug}$}
    Fine-tune without augmentation $\mathcal M $ = \textit{fine-tune}$(PLM, D)$ \;
    \For{$x_i$ in $\{x_i\}_{i=1}^{n}$}{
        Get the $Matches_i$ in $x_i$ by Algorithm~\ref{algorithm:localization}\;
        Generate augmented samples $D^{aug}_{i}$ with $\mathcal{G}$ following the steps in Figure~\ref{fig:augmentation}\;
        Initialize the prediction probabilities of $D^{aug}_{i}$ as $P^{aug}_{i}$\;
        \For{$x_i^j$ in $D^{aug}_{i}$}{
            Calculate the prediction probability $p_{i}^{j} = prob(\mathcal{M}(x_{i}^{j}) = y_i)$\;
            Add $p_{i}^{j}$ to $P^{aug}_{i}$\;
        }
        Calculate the sampling weights of $D^{aug}_{i}$ according to Eq.~\ref{eq:sampling} \;
        Sample 5 samples from $D^{aug}_{i}$ by weights and add them to $D^{aug}$\;
    }
    \Return $D^{aug}$\;
\end{algorithm}

\subsection{KGs Preprocessing}
\label{sec:kg_preprocess}

Healthcare is a field with rich professional knowledge. There are also publicly available knowledge graphs, e.g., the Unified Medical Language System (UMLS) \cite{DBLP:journals/nar/Bodenreider04}. We take such open medical KGs for healthcare \method. UMLS Metathesaurus is a compendium of many biomedical terminologies with the associated information, including synonyms, categories, and relationships.
It groups semantically equivalent or similar words into the same concept, for example, the words `flu', `syndrome flu' and `influenza' are mapped to the same concept unique identifier (CUI)  \textit{C0021400}, which belongs to the category, \textit{disease or syndrome}.
There are 127 semantic types in biology, chemistry, and medicine, consisting of 4,441,326 CUIs (16,132,273 terminologies) in the UMLS 2021AA version.
Since the size of the KG is too large to affect the speed of retrieval, we only screen out entities that belong to the type of medicine (e.g., \textit{body part, organ, or organ component}, \textit{disease or syndrome}, etc.). 
At the same time, we also delete non-English strings.
Finally, we keep 1,145,062 CUIs (16 semantic types), 502 types of relationships and 4,884,494 triples.
Although there are Chinese medical terminologies in UMLS, the number is limited. 
Hence, we use an open-source Chinese medical KG, CMedicalKG,\footnote{https://github.com/liuhuanyong/QASystemOnMedicalKG} which includes 44,111 entities (7 categories), 10 types of relationships, and 294,149 triples.

\section{Case Study}

\begin{figure}[t]
    \centering
    \includegraphics[width=1.0\linewidth]{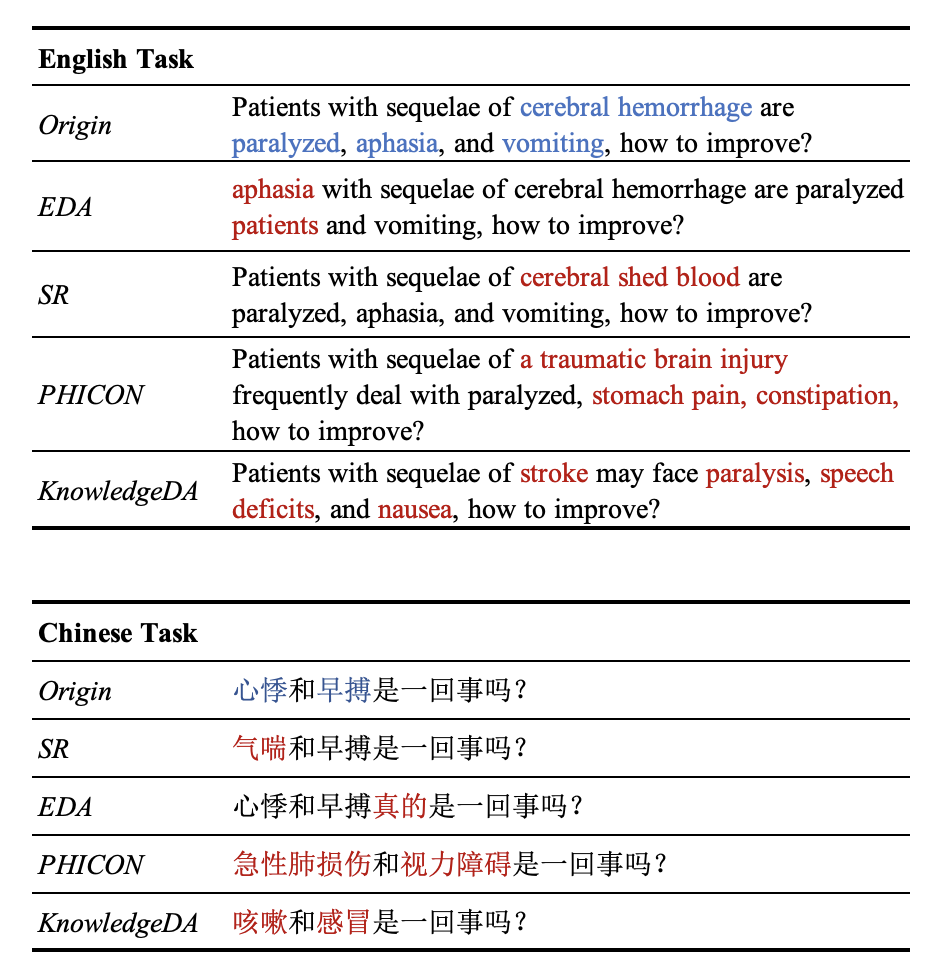}
    \vspace{-1.5em}
    \caption{Examples of data augmentation in English and Chinese. Text chunks in {\color{blue}blue} are the entities in the original sentence and text chunks in {\color{red}red} are the modified words/entities by DA methods.}
    \label{fig:case_study}
    \vspace{-0.5em}
\end{figure}

Fig.~\ref{fig:case_study} shows the examples in English and Chinese with various DA methods. 
We can observe that the sentence augmented by \method has a high quality as it can introduce more domain entities and the whole sentence has a good semantic meaning.

\section{Text Classification in Software Development}
\label{sec:result_software}

\subsection{Dataset} 
We use a open data, \textit{SO-PLC}\footnote{\scriptsize{http://storage.googleapis.com/download.tensorflow.org/data/stack\_\\overflow\_16k.tar.gz}}, which is
a Stack Overflow dataset for 4 programming language classification: python, C\#, java, and javascript.

\subsection{KG for Software Development}
There is little research on building KGs for software development NLP tasks, and thus we decide to build one from scratch.

To build the KG, we refer to the software developer forum \textit{Stack Overflow} to obtain raw text data.\footnote{\scriptsize{https://stackoverflow.com/}}
\textit{Stack Overflow} is one of the biggest forums for professional and enthusiastic software developers. 
Various technical questions are covered on the platform and marked with appropriate tags. 
These tags are usually programming-specific terminologies and can be beneficial to learn about tech ecosystems and the relationships between technologies \cite{doi:10.1080/09720529.2020.1721857}.
To build a KG from tags, we follow the existing KG construction process \cite{li2020real}:

\textit{Step 1. Data Collection}: We use programming languages (e.g., python, C\#, java, and javascript) as keywords to search for related questions on Stack Overflow, and sort them according to `most frequency'; 
then crawl the tags that appeared in the top 7,500 related questions (i.e., the first 150 pages).

\textit{Step 2. Entity Recognition}: A tag is a word or phrase that mainly describes the key information of the question, which is usually a programming-specific terminology \cite{doi:10.1080/09720529.2020.1721857}.
Hence, we directly treat tags as the entity names in the KG.

\textit{Step 3. Relation Formation}: There is usually more than one tag in one question. When multiple tags co-appear at the same question, we link them in the KG.
Afterward, there is still a lack of entity types and edge types, and we use the community detection algorithm, Louvain \cite{Blondel_2008}, to automatically classify tags, and the edge type is defined by the types of the two connected entities.

Finally, we get TagKG, which includes 6,126 entities (11 categories), 56 types of relationships, and 41,227 triples.

\begin{table}[t]
    \centering
    \scriptsize
    \begin{tabular}{lcc}
    \toprule
    \multicolumn{1}{c}{\multirow{2}{*}{\textbf{Method}}} & \multicolumn{2}{c}{\textbf{SO-PLC}} \\
    \cmidrule(r){2-3}
    & Acc. & F1 \\ \hline
    \textit{None} & 84.78(0.48) & 84.65(0.50) \\
    \textit{SR} & 84.72(0.32) & 84.69(0.31) \\
    \textit{EDA} & 84.78(1.08) & 84.71(1.08) \\
    \textit{PHICON} & 85.63(1.17) & 85.60(1.19) \\
    \textit{\method} & \textbf{86.82(0.90)*} & \textbf{86.83(0.88)**} \\ \bottomrule
    \end{tabular}
    \vspace{-.5em}
    \caption{\small{Performance on SO-PLC dataset (\textit{BERT} as PLM)}}
    \label{tab:MainRes_Program}
    \vspace{-2em}
\end{table}

\subsection{Result} 
As illustrated in Table \ref{tab:MainRes_Program}, there are almost no improvements or even slight decreases with EDA and SR, meaning these general DA methods are not suitable for the texts in software development forums.
With the help of our constructed TagKG, \textbf{PHICON} achieves some performance gains by replacing same-category programming entities; this indicates that the category identified by the community detection algorithm in TagKG is effective for understanding software development related texts.
By leveraging TagKG more comprehensively, \method works even better and improves accuracy and F1 score by 2.42\% and 2.58\%, respectively, compared with no-augmentation. 
It also implies that the construction of TagKG is valid.

\end{document}